# Exploiting Facial Landmarks for Emotion Recognition in the Wild

Matthew Day
Department of Electronics, University of York, UK
matt.day@york.ac.uk

## ABSTRACT
In this paper, we describe an entry to the third Emotion Recognition in the Wild Challenge, EmotiW2015. We detail the associated experiments and show that, through more accurately locating the facial landmarks, and considering only the distances between them, we can achieve a surprising level of performance. The resulting system is not only more accurate than the challenge baseline, but also much simpler.

## Categories and Subject Descriptors[1]
I.2.10 [Artificial Intelligence]: Vision and Scene Understanding – modeling and recovery of physical attributes, shape, texture.

## 1. INTRODUCTION
Accurate machine analysis of human facial expression is important in an increasing number of applications across a growing number of fields. Human computer interaction (HCI) is one obvious example. Others include medical monitoring, psychological condition analysis, or as a means of acquiring commercially valuable feedback.[1]

Whilst the problem was traditionally addressed in highly constrained scenarios, an increasing focus on 'in the wild' (i.e. unconstrained) data has emerged in recent years. In this respect, the EmotiW2015 challenge [5] aims to advance the state of the art. The challenge is divided into two sub-challenges: (1) audio-video based and (2) static image based. Audio, in particular, has been shown to contain discriminative information [4] and motion intuitively provides valuable cues to a human observer. However, effectively exploiting this information is undoubtedly challenging. This fact is demonstrated by the two baseline systems [5], which achieve negligible accuracy difference across the sub-challenges. It is then clearly a useful pursuit to consider only static images, because accuracy improvements here can be built on in more complex systems that analyze video. Consequently, this work focusses on the image based sub-challenge.

---

[1] This paper was originally accepted to the ACM International Conference on Multimodal Interaction (ICMI 2015), Seattle, USA, Nov 2015. It has been made available through arXiv.org because the author was unable to present.

All images in this sub-challenge contain an expressive face with the goal to assign an emotion label from the set {neutral, happy, sad, angry, surprised, fearful, disgusted}. These labels originate from the work of Ekman [7], who noted that facial expressions are primarily generated by the contraction or relaxation of facial muscles. This causes a change in the location of points on the face surface (i.e. facial landmarks). Whilst other cues may exist, such as coloring of the skin or the presence of sweat or tears, shape changes remain the most significant indicator.

The relationship between muscle movements and emotion has been well studied and is defined by the emotional facial action coding system [9] (EMFACS). For example, happiness is represented by the combination of 'cheek raiser' with 'lip corner puller'. Sadness is demonstrated by 'inner brow raiser' plus 'brow lowerer' together with 'lip corner depresser'. However, in the static image sub-challenge, it is not possible to detect movements, so how well can expression be predicted from a single image? In contrast to the EmotiW2015 challenge, most prior-art has reported results on well-lit well-aligned faces. For a 5-class problem, with fairly accurate registration, [1] demonstrated classification accuracies of around 60%.

Section 2 of this paper discusses face detection and landmark location. Sections 3 and 4 describe the features and modelling approaches used in our experiments. Section 5 provides the main experimental results and section 6 discusses what we can take from these as well as offering some miscellaneous considerations.

## 2. FACE REGISTRATION
The first stage in any system of facial analysis involves locating and aligning the face. Methods which holistically combine locating a face with locating facial landmark points have clear appeal. In particular, deformable parts models (DPM), introduced in [8] have become one of the most popular approaches in the research community.

On the other hand, face detection and landmark location have both been extensively studied separately. Many excellent solutions have been proposed to both problems and our own experience suggests that tackling the tasks separately may have advantages in some scenarios. In particular, a recent method [12] for facial landmark location has excellent performance on unconstrained images and is therefore well suited to the EmotiW2015 challenge. The method uses a sequence of gradient boosted regression models, where each stage refines the position estimate according to the results of many point-intensity comparisons. We use the implementation of [12] provided by the dlib library [13]. The model, provided for use with this library, was trained using the data from the iBUG 300-W dataset and it positions 68 points on frontal faces, similar to the baseline.

In [15], face detection based on rigid templates, similar to the classic method of [18], achieves comparable accuracy to a detector based on DPM [8], but the former has a substantial speed advantage. We choose the rigid template detector included in the dlib library [13] as a result of informal performance comparisons. This method uses histogram of oriented gradients [3] (HoG) features combined with a linear classifier. It has a very low false positive rate, although it fails to find a face in almost 12% of the challenge images. In these cases, we roughly position a bounding square manually to allow subsequent processing.[2]

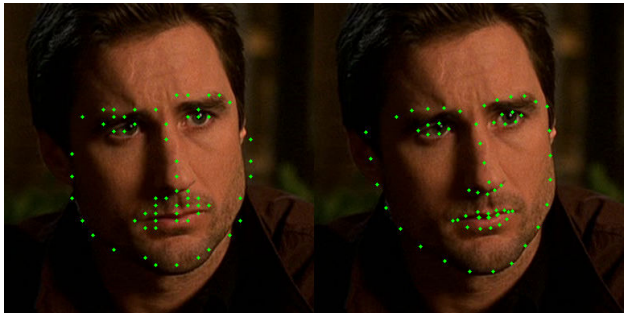

**Figure 1: Example landmarks from baseline (left) and proposed system (right)**

Figure 1 shows a representative comparison of the landmark points output by our system and those of the baseline system. To try to quantify the advantage here, we inspect the automatically located points for each image in the training set and give each one a subjective score based on how well they fit the face. In Figure 1 for example, we would consider the baseline points as 'close' and our points as 'very close'. The results from this exercise are shown in Table 1. It is clear from this that we have a better starting point for using shape information to estimate emotion – experiments in later sections quantify this further.

**Table 1. Accuracy of baseline points versus proposed system**

|  | Excluded | Fail | Poor | Close | Very Close |
|---|---|---|---|---|---|
| Baseline | 67 | 55 | 173 | 663 | 0 |
| Proposed | 0 | 2 | 37 | 215 | 704 |

## 3. FEATURES

Our final system, i.e. challenge entry, uses only one very simple type of shape feature. However, we performed experiments with various features which we describe in the following paragraphs.

## 3.1 Shape

Intuitively, given accurate facial landmark locations, we can infer a lot about facial expression – arguably more than is possible from only texture. We consider two simple types of shape feature, both derived from automatically located facial landmark locations. For both types, we first normalize the size of the face.

---

[2] After our main experiments were complete, we found a combination of open source detectors could reduce the miss rate to 3% and the landmark estimator is unlikely to perform well on the remaining difficult faces, regardless of initialization.

### 3.1.1 Distances between Points
We consider the distances between all distinct pairs of landmark points. We have 68 landmarks, giving 2278 unique pairs. Many of these pairs will contain no useful shape information and we could add heuristics to reduce this number considerably, although this is not necessary for the models we subsequently learn.

### 3.1.2 Axis Distances from Average
We speculate that the point-distances may not capture all shape information alone. We therefore test a second type of feature that considers the displacement from the average landmark location, where the average is taken from all faces in the training set. After up-righting the face, for each point, we take the *x*- and *y*-distances from the average location as feature values. This results in a vector of length 136.

## 3.2 Texture
By including texture to complement the shape information, we hope to improve classification accuracy in our experiments. We note that the baseline system [5] is based entirely on texture features and many previous successful approaches have also used texture, e.g. [1].

### 3.2.1 Biologically Inspired Features (BIF)
BIF [10] are perhaps most well-known for the success they have achieved in facial age estimation. However, they have also demonstrated excellent performance in other face processing problems [16] and have been applied to the classification of facial expressions [14]. As a rich texture descriptor, based on a model of the human visual system, BIF would appear to represent a good candidate for this application.

Evaluation of BIF involves applying a bank of Gabor filters with different orientations and scales to each location in the face image. The responses of these filters are pooled over similar locations and scales via non-linear operators, maximum (MAX) or standard-deviation (STDDEV). In practice, the aligned face image is partitioned into overlapping rectangular regions for pooling and the pooling operation introduces some tolerance to misalignment.

Our implementation closely follows the description in [10]. We extract 60x60 face regions, aligned according to the automatically located landmarks. We use both MAX and STDDEV pooling, with 8 orientations and 8 bands. Our implementation then has 8640 feature values.

### 3.2.2 Point Texture Features
We speculate that we may gain more information from texture features that are more directly tied to the location of landmarks. We therefore also consider a second type, where the feature values are simply Gabor filter responses at different sizes and orientations, centered on each landmark location. We refer to these as 'point-texture features'. We evaluate filters at 8 scales and 12 orientations, giving a total of 6528 feature values for the 68 landmark points.

## 4. MODELLING
To construct predictive models using our features, we use two standard approaches from the machine learning literature: support vector machines (SVM) [11] and gradient boosting (GB) [11]. For the SVM classifiers, we use the implementation provided by libsvm [2] with a RBF kernel. We optimize the *C* and *gamma* parameters on the validation data via a grid search, as advocated by the authors of [2]. For the GB classifiers, we use our own implementation. We find that trees with two splits and a shrinkage factor of 0.1 generally work well on this problem, so we fix these parameters and optimize only the number of trees on the validation data.

## 5. EXPERIMENTS

In all of the following experiments, only the challenge training data are used to construct the model, with parameters optimized on the validation set. At one point, we experimented with combining the training and validation data and learning using this larger set. However, this did not result in an improvement in accuracy on the test data, so we did not pursue this approach or include the result.

### 5.1 Simple Shape-based Classifiers

We start using only the point-distance features described in 3.1.1. We learn SVM and GB classifiers which give the performance figures shown in Table 2.

**Table 2: Main performance figures**

| Model | Train | Validate | Test |
|---|---|---|---|
| GB | 60.1% | 40.8% | 44.4% |
| SVM | 52.1% | 37.4% | **46.8%** |
| Baseline | - | 36.0% | 39.1% |

A confusion matrix for the SVM classifier on the test data is shown in Table 3.

**Table 3: Test data confusion matrix for challenge entry**

| Estimate → / Truth ↓ | Angry | Disgust | Fear | Happy | Neutral | Sad | Surprise |
|---|---|---|---|---|---|---|---|
| Angry | 29 | 1 | 4 | 5 | 10 | 7 | 13 |
| Disgust | 3 | 0 | 0 | 6 | 4 | 4 | 0 |
| Fear | 13 | 0 | 1 | 3 | 13 | 6 | 5 |
| Happy | 5 | 0 | 0 | 67 | 7 | 16 | 0 |
| Neutral | 3 | 0 | 1 | 3 | 40 | 9 | 2 |
| Sad | 9 | 0 | 5 | 5 | 12 | 16 | 8 |
| Surprise | 8 | 0 | 2 | 0 | 6 | 0 | 21 |

### 5.2 Classifiers using Other Features

Taking each of the other three types of feature described in section 3 in turn, we add to the point-distance features. Surprisingly, in each case, we did not observe any improvement on validation data over using the point-distance features alone.

For the features of 3.1.2, the accuracy on validation data actually dropped slightly. This could be a result of using a slightly different procedure for size normalization with these features. However, there is also a concern that the average point locations were not useful, due to the large variations in pose.

For the texture features of 3.2, the result was more surprising. We expected these to add some useful information, but this appeared not to be the case, despite their quantity far exceeding that of the distance features.

As a consequence, we conclude that the simple point-distance features already contain the most information relevant to the task. We use the SVM model from Table 2 as our challenge entry.

### 5.3 Improvement over Baseline

To quantify the advantage that our more accurate landmark locations bring over the baseline, we learn directly comparable models using both sets of points. For the baseline system, landmark points are not available for all images, because the face detector fails in some cases. For a fair comparison, we therefore use exactly the same subset of images across train/validation/test sets in both trials. Where no points exist for a test image, we assign a 'Neutral' label.

Table 4 shows the results of this comparison. From these we can conclude that the landmarks used in the proposed system provide a very clear advantage over those of the baseline system.

**Table 4: Overall accuracy using baseline points and points from proposed system**

| Landmarks | Model | Train | Validate | Test |
|---|---|---|---|---|
| Proposed | GB | 65.8% | 40.4% | 38.4% |
| Proposed | SVM | 50.0% | 41.2% | 40.6% |
| Baseline | GB | 53.9% | 32.3% | 31.2% |
| Baseline | SVM | 47.9% | 34.1% | 27.7% |

## 6. DISCUSSION

Considering the results of Table 3, performance on each class of emotion exhibits the same pattern seen in previous EmotiW challenges. Specifically, performance is promising for faces with neutral (SVM:69%,GB:64%), happy (71%,62%), angry (42%,46%), and surprise (57%,43%) expressions. On the other hand, sad (29%,25%) and fearful (2%,17%) expressions are more difficult to distinguish. The subtleties of disgust (0%,0%) might be impossible to detect using such simple features taken from static images. Indeed, this task is not only difficult for machines, but without contextual information it is also difficult for humans to distinguish disgust from other more prevalent emotions. Our overall accuracy is more than three times better than random guessing, representing a small improvement over the accuracy achieved in [1] on more constrained static images. The final system we propose achieves 47% accuracy on the test data, whilst the baseline achieves 39% accuracy.

Comparing our SVM and GB classifiers, the former lead to slightly better results in most cases, whereas the latter are significantly simpler and faster to evaluate. However, model complexity differences become insignificant if texture features must be evaluated as this dominates time required to evaluate either type. The key advantage of our proposed system is that the distance features are trivial to evaluate in comparison to commonly used features such as BIF [10], LBP [17] or HoG [3].

The GB model allows the influence of its features to be examined and Figure 2 is a result of this analysis. The distance from the eyes to the corners of the mouth clearly has the greatest influence. This seems reasonable considering the degree to which a mouth is upturned or downturned is one of the clearest indicators of emotional state. Figure 2 also includes distances indicative of eye and mouth openings, which are also intuitively discriminative.

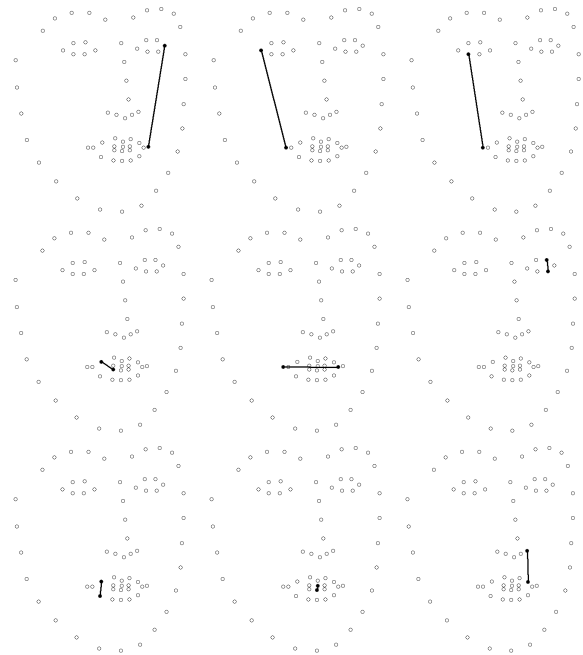

**Figure 2: From left to right, then top to bottom, the most influential distances in our gradient boosted model**

Before concluding, we must note an observation that potentially affects the baseline accuracy. Almost all of the challenge images have an incorrect aspect ratio that results in elongated faces. We manually correct this prior to performing our experiments. If we instead use the images as provided, the face detector finds only around 60% of faces. Given that we are particularly interested in modelling shape here, it is important to work with consistent aspect ratios.

As a final comment, although the landmarks found by our system are more accurate than those in the baseline, there is still much scope for improvement. Given 100% accurate landmark locations, an interesting line of further work might be to tailor the modelling approach to the problem in an attempt to see just how far static shape alone can be used in estimating facial expression.

## 7. ACKNOWLEDGMENTS

Our thanks to Professor John A. Robinson for his support in this work.